\documentclass[letterpaper, 10 pt, conference]{ieeeconf}  

\usepackage{graphicx,caption}
\usepackage{array}
\usepackage{tabularx}
\usepackage{multirow}
\usepackage{amssymb}
\usepackage{amsmath} 
\IEEEoverridecommandlockouts
\usepackage[dvipsnames]{xcolor}

\title{\LARGE \bf
Trust-Aware Emergency Response for A Resilient Human-Swarm Cooperative System
}

\author{Yijiang Pang$^{}$ and Rui Liu$^{*}$
\thanks{* is the corresponding author {\tt\small ruiliu.robotics@gmail.com}.}
\thanks{$^{}$ Authors are with The Cognitive Robotics and AI Lab (CRAI), College of Aeronautics and Engineering, Kent State University, Kent, OH 44240, USA.}
}

\begin{document}

\maketitle
\thispagestyle{empty}
\pagestyle{empty}

\begin{abstract}
A human-swarm cooperative system, which mixes multiple robots and a human supervisor to form a mission team, has been widely used for emergent scenarios such as criminal tracking and victim assistance. These scenarios are related to human safety and require a robot team to quickly transit from the current undergoing task into the new emergent task. This sudden mission change brings difficulty in robot motion adjustment and increases the risk of performance degradation of the swarm. Trust in human-human collaboration reflects a general expectation of the collaboration; based on the trust humans mutually adjust their behaviors for better teamwork. Inspired by this, in this research, a trust-aware reflective control (\textit{Trust-R}), was developed for a robot swarm to understand the collaborative mission and calibrate its motions accordingly for better emergency response. Typical emergent tasks ``transit between area inspection tasks", ``response to emergent target -- car accident" in social security with eight fault-related situations were designed to simulate robot deployments. A human user study with 50 volunteers was conducted to model trust and assess swarm performance. Trust-R's effectiveness in supporting a robot team for emergency response was validated by improved task performance and increased trust scores.
\end{abstract}


\section{INTRODUCTION}
A human-swarm cooperative system, which mixes multiple robots and a human supervisor to form a mission team, has been widely used for complex and large-scale scenarios, such as unmanned aerial vehicle (UAV) swarm cooperated navigation in an unknown environment \cite{McGuire2019Minimal, Soares2018Group}, UAV team assisted precision agriculture \cite{Albani2017agricultural, Albani2017weed}, and mixed swarm-ground-vehicle team for forest inspection and wildlife protection \cite{Carpentiero2017wheeled, Duarte2016marine}. Many scenarios include emergent situations such as immediately switching to track a criminal vehicle during city patrolling and switching to a more urgent disaster area for victim rescue. These scenarios involve human safety and impose strict requirements on the response time and mission quality towards a robot team. Therefore, upon receiving assistance requests from an emergent scenario, a robot team is expected to quickly terminate the current task and transit into the new emergent one, bringing difficulty in motion adjustment for the swarm. In these emergent transitions, uncertainties such as position error and direction deviation, are accumulated from prior tasks, influencing swarm performance and even causing robot failures and further cooperation failure. 
Moreover, real-world abnormalities such as motor degradation, sensor failure, and wind disturbances further degrade swarm performance during emergency response.
Without real-time estimation and adjustment of the performance, it is challenging for a robot swarm to respond to emergent tasks with the human desired quality.

\begin{figure}[t]
  \centering
  \includegraphics[width=0.9 \linewidth ]{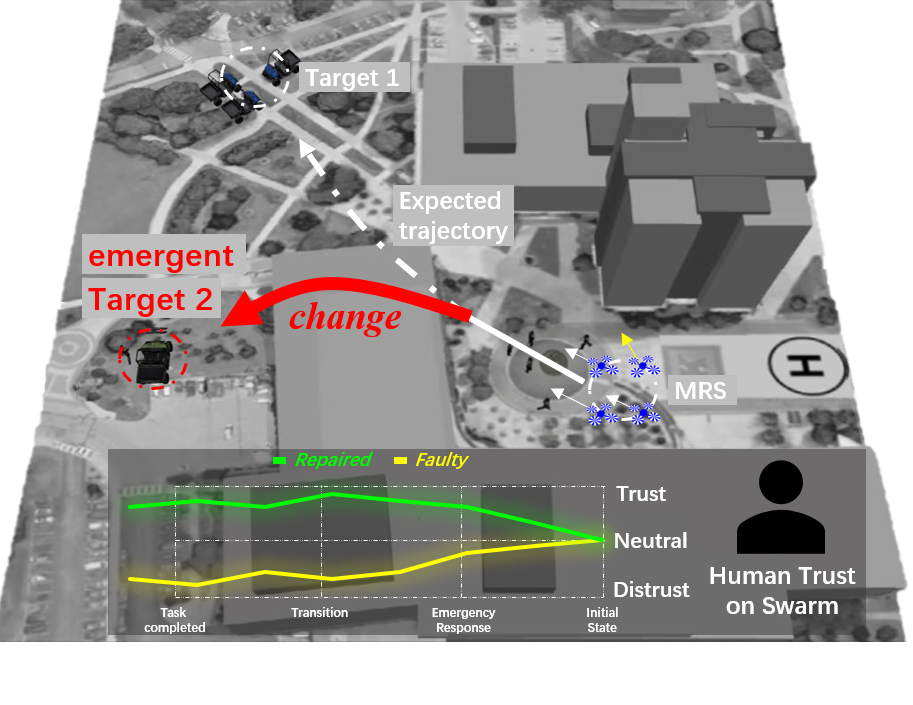}
  \caption{Illustration of the trust-aware reflective control of a swarm for emergency response. When a robot swarm appears faulty behaviors caused by faulty robots appear in a robot swarm, the \textit{\textbf{Trust-R}} corrects the swarm's faulty behaviors to regain human trust.}
  \label{illustration_1}
  \vspace{-1em}
\end{figure}

Trust in human-human collaboration reflects a general expectation of the collaboration; based on the trust humans mutually adjust their behaviors for better teamwork. Inspired by this human-human trust, the human-swarm trust was modeled to estimate human expectation on swarm behaviors. A higher trust will lead to a higher autonomy level, while a lower trust will trigger robot behavior corrections. Specifically, this research developed a trust-aware reflective control \textit{\textbf{Trust-R}} method to update swarm behaviors based on the estimated trust level of robots, maintaining a high-quality human-supervisory emergency response. 

As shown in Figure 1, supported by a trust estimation, a weighted update algorithm helps a robot to increase information sharing with its trusted robot neighbors and decrease information sharing with distrusted neighbors, constraining negative influences from abnormal robots onto the whole swarm. In this way, accumulated errors are mitigated and faulty behaviors are corrected in an early stage to ensure a high-quality collaboration. In this paper, there are mainly three contributions.

\begin{itemize}
    \item 
    A trust-aware cognitive control algorithm, \textit{\textbf{Trust-R}}, has been developed based on control laws and trust psychology, to enable a robot to selectively share information with its team members based on the estimation of trustworthiness of others.
    
    \item
    A self behavior reflection mechanism has been developed for cooperation calibration between a swarm and a human. Based on the trust estimation, \textit{\textbf{Trust-R}} enables a swarm to self-diagnose and identify the unsatisfied behaviors, providing in-process calibration for better human-swarm collaboration.
    
    \item
    A novel research framework of ``behavior-repair to trust-repair" has been developed and validated, theoretically proving that robot behavior corrections can repair human trust and further can calibrate human-swarm collaboration. It provides a guideline for future research of autonomy-related trust repair and maintenance.
    
\end{itemize}

\begin{figure*}[h!]
  \centering
 \includegraphics [width=0.9 \linewidth ]{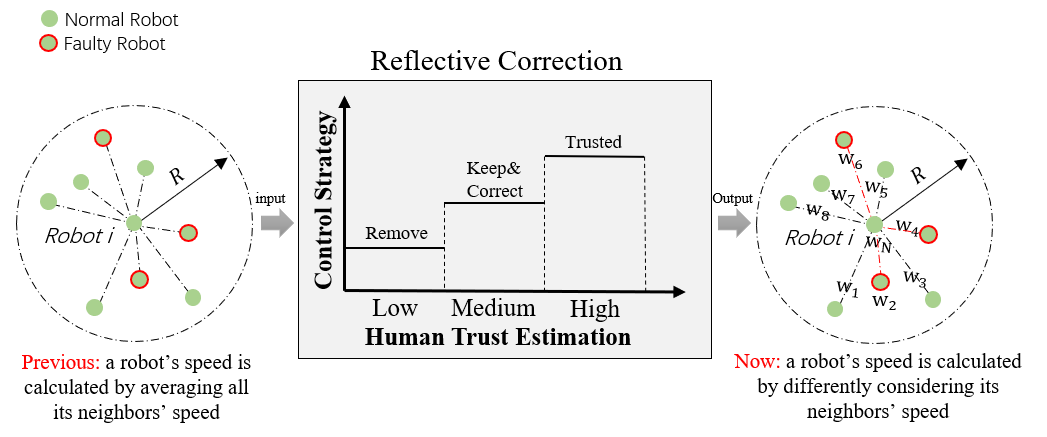}
  \caption{Trust-aware reflective control for emergency response in human-swarm system deployment. The weighted communication quality method enhances the information exchanging and connectivity between trustworthy robots and decreases information sharing between faulty robots.}
  \label{illustration_2}
  \vspace{-1em}
\end{figure*}

\section{Related Work}
Trust repair mechanisms were investigated previously.
Robinette et al. and Schweitzer et al. evaluated methods of repairing trust in human-robot cooperation through robot apology of its mistakes or promise of improved performances \cite{Robinette2015Timing, Schweitzer2006Promises}.
Wang, et al. investigated the explanation mechanism's impact on trust repair to help a robot restore human trust by explaining the decision-making processes \cite{Wang2016calibration}.
However, due to the experiment settings of assessing trust after tasks, the study did not explore the in-process changes of the trust, which is critical for a continuous human-robot interaction; these trust models are fragile that when cooperation performance is poor the established trust will be broken, making it difficult to support reliable cooperation with dynamic interactions.
On the other hand, this paper focuses on trust repairing for the situations that when trust is damaged by unstable robot performance. Repairing human trust on a swarm is critical for maintaining a high-quality collaboration given trust decides willingness and expectation of the cooperation.

Prior research investigated fault detection in dynamic or emergent swarm deployment \cite{Visinsky1994fault, Khaldi2017Monitoring}.
Tarapore et al. developed behavior-based approaches to distinguish normal robots from abnormal ones by identifying the inconsistent or rare robot behaviors under dynamic transitions \cite{Tarapore2017fault, Tarapore2019Fault}. Christensen et al. developed a general abnormality detection approach with dynamic task setting to detect non-operational robots in a robot swarm by identifying different flashing frequency of each robot's on-board light-emitting diodes \cite{Christensen2009fireflies}.
However, it is challenging for robots to find out higher-level faulty issues through self-diagnosis without a holistic view, such as an ant mill.
This paper investigated a fault detection method for the robot swarm by developing a robot understanding of human trust. According to an estimation of human trust in each robot in swarm, \textit{\textbf{Trust-R}} helps a robot swarm to proactively self-correct its faulty behaviors.

In our previous work \cite{liu2019trust}, a decentralized trust-aware behavior reflection method \textit{\textbf{Trust-R}} was developed to correct faulty behaviors of a swarm. The previous version method was initially developed to only correct abnormal heading directions of a swarm. This paper further validated the method's effectiveness in supporting emergency response, where immediate disengaging an undergoing task causes abnormal robot behaviors related to speed, motion direction and team spatial distribution, further undermining swarm performance. Moreover, real-gravity simulations with physical UAV models were adopted to explore a more applicable control method for swarm emergency response.

\section{Distrusted Flocking in Emergency Response}
\subsection{Illustrative Scenario for Swarm Correction}
Consider a robot swarm of $n$ holonomic robots with positions $X_i\in \mathbb{R}^3$, where $X_i = (\boldsymbol{x}_{i,h},\boldsymbol{x}_{i,v},\theta_{i})$. 
Every robot $i$ only communicates with its direct neighbors $j \in N_i$, where $N_i$ is the set of all neighbors of $i$ within the communication radius, $R$. 
A robot $i$ is controlled by the linear velocity $\boldsymbol{u}_{i}^{v}$ and angular velocity $\boldsymbol{u}_{i}^{w}$ generated by motors. $\boldsymbol{x}_{i,h}$, $\boldsymbol{x}_{i,v}$, and ${\theta}_{i}$ denotes horizontal position, vertical position, and orientation state, respectively. 
The dynamic model for each robot is detailed in our previous paper \cite{liu2019trust}.

At each time step $t$, a robot $i$ updates its motion status by averaging its neighbors{'} motion status,
\begin{equation}\label{eq:10} \boldsymbol{u}_i[t+1]=\frac{1}{N_i+1}(\boldsymbol{u}_i[t]+\sum_{j\in N_i}^{} \boldsymbol{u}_j[t]). \end{equation}

As seen from the distributed update method above, faulty robots will relay unreliable motion information to their neighbors, which in turn will mislead their neighbors’ motions.

\subsection{Emergency Response with Accumulated Uncertainty}
With a typical setting in a hierarchical swarm control \cite{gupta2015survey}, the robots in the swarm are usually divided into two types: leader robots and follower robots, and only leader robots can receive control information from the base station, such as dynamic destination coordinates and cruising speed.
Meanwhile, faulty robots accumulate motion uncertainty, such as location shifting, heading direction deviation, and extra speed, mapping into the swarm because of the update of consensus policy. In this paper, given the speed and heading direction requirement in the flocking, the accumulated uncertainty is described by the accumulated speed $\delta{\boldsymbol{u}}$ and accumulated location shifting $\delta{\boldsymbol{x}}$ of the swarm compared with the expected swarm status of receiving the new-task assignment. Given that assumption, the velocity $\boldsymbol{u}_i$ of the leader robots is updated using equation \ref{eq:11},

\begin{equation}
\label{eq:11}
    \begin{aligned}
    \boldsymbol{u}_i[t+1]=
    \frac{1}{N_i+1}(\boldsymbol{u}_i[t]+\sum_{j\in N_i}^{} \boldsymbol{u}_j[t]) + \boldsymbol{u}^\gamma_i[t]
    \end{aligned},
\end{equation}
where ${u}^\gamma_i$ is the navigational feedback and accumulated uncertainty and is given by
\begin{align*} 
\boldsymbol{u}^\gamma_i[t]&:={f}^\gamma_i({x}_{i}[t], {x}_\gamma, \delta{x}, {u}_{i}[t], {u}_\gamma, \delta{u})\\
&=-c^\gamma_1(\boldsymbol{x}_{i}[t]-\boldsymbol{x}_\gamma+\delta{\boldsymbol{x}})-c^\gamma_2(\boldsymbol{u}_{i}[t]-\boldsymbol{u}_\gamma+\delta{\boldsymbol{u}})
\end{align*}
, and the $\gamma$ - robot $(\boldsymbol{x}_\gamma, \boldsymbol{u}_\gamma)$ is the virtual leader that leads the swarm to follow its trajectory \cite{La2009Adaptive}. The parameters $\boldsymbol{x}_\gamma$ and $\boldsymbol{u}_\gamma$ are the destination and cruising speed that the leader robots get from the base station. $c^\gamma$ denotes the gain of the components.
During a normal task execution, $|{x}_{i}[t]-{x}_\gamma|\rightarrow0$, $|{u}_{i}[t]-{u}_\gamma|\rightarrow0$, as $t\rightarrow\infty$ \cite{Ren2007consensus}. When the swarm responds to an emergent task, the expected swarm status of receiving the new-task assignment will be influenced by $\delta{x}$, $\delta{u}$.
As seen from the dynamic task update method above, faulty robots will be able to introduce uncertainty to the swarm and worsen the swarm task performance. In order to mitigate the uncertainty for swarm performance assurance which is important to ensure a high-quality emergent task response, our method will target on suppressing negative influence $\delta{x}$, $\delta{u}$ in real time, to finally correct swarm behaviors.

\section{Trust-Aware Reflective Control for Emergency Response}
The architecture of the \textit{\textbf{Trust-R}} is shown in Figure 2. 
With \textit{\textbf{Trust-R}}, an understanding of human-expected behaviors is developed to determine the robot's communication quality with its neighbors, maximally reducing the negative influence of a faulty robot on the whole swarm.

\subsection{Human Supervision}

In human-swarm cooperation, the human serves as the operator to monitor and guide task execution of the robot swarm.
The operator is expected to distinguish between current performance and expected performance, identify robots implicated with fault, and rate the current performance for individual robots and the whole swarm with trust level (e.g., Low, Medium, and High), as shown in Figure \ref{illustration_2}.
The proposed method, \textit{\textbf{Trust-R}}, is semi-automated, which integrates the human's estimated trust level for considering robots differentially.

\subsection{Trust-Aware Connectivity}
In general, each robot in the swarm calculates its speed by averaging all its neighbors’ speed to reach consensus. However, when faulty robots appear in a swarm, they have a negative influence on swarm performance, bringing the risk of failure to assigned emergent tasks. 
This paper used a weighted connection method, the weighted mean subsequence reduced algorithm \cite{Saldana2017Resilient}, to update information based on robot trustworthiness. The velocity of each robot $u_i$ is updated with weighted reference to its neighbors,

\begin{equation}
\label{eq:12}
    \begin{aligned}
    u_{i}[t+1]=w_{i}[t]u_{i}[t]+\sum_{j\in{N_i}}w_{j}[t]u_{j}[t].
    \end{aligned}
\end{equation}

\subsection{Trust-Aware Communication Quality Assessment}
The overall communication graph for robot $i$ is $\boldsymbol{\mathcal{E}} = \{(i, j) \mid j\in N_i\}$. Based on the estimated trust levels of the two robots $\{i, j\}$, communication quality, $f_{ij}\in [0,1]$, is used to measure the reliability of exchanged information. The trust-aware communication quality is dynamically updated to reflect the changing communication graph using Equation \ref{eq:13}. The best communication distance between two robots $i$ and $j$ is $\rho$. Communication within $\rho$ is considered as communication with the best quality. The parameter, $\eta$, is used as a weighting factor to discourage the impact of faulty robots on their neighbors.

\begin{equation}
\label{eq:13}
    f_{ij}=\left\{
                \begin{array}{ll}
                  0 & ||\boldsymbol{x}_i-\boldsymbol{x}_j|| \geq R\\
                  \frac{1}{2}(g_i+g_j)\eta & ||\boldsymbol{x}_i-\boldsymbol{x}_j||\leq \rho \\ 
                \frac{(g_i+g_j)\eta}{2}\exp{\frac{-\gamma (||\boldsymbol{x}_i-\boldsymbol{x}_j||-\rho)}{R-\rho}} & otherwise \\
                
                \end{array}
              \right.,
\end{equation}

\noindent where $g_i$ is the trust level of robot $i$, which is estimated based on a human user study. Within the communication range, the communication reliability is decided by trust values of the two involved robots.

The rationale for designing the trust-aware communication quality is to encourage information sharing with trusted robots by using higher limits to allow more information to flow, while discourage information sharing with untrusted robots by using lower limits to reduce the information flow, gradually isolating the faulty robots from the normal ones. 


\subsection{Trust-Aware Behavior Correction}
A swarm proactively corrects its faulty behaviors with the following process. \textit{\textbf{Trust-R}} encourages the information sharing between trusted robots, gradually making the robot behaviors consistent with the most trusted ones. While the method restrains the information sharing from faulty robots to trusted robots, gradually isolating the faulty robots from other robots and even abandon the faulty robots from the team when the fault is severe. The trust levels on robot behaviors are estimated based on our pioneer human trust study.


\begin{equation}
\label{eq:17}
w_k[t]=\frac{\hat{f}_k [t]}{\hat{f}_i [t]+\sum_{j \in N_{i}}\hat{f}_j[t]}, k\in \left [ i,N_i \right ]
\end{equation}

Weights for updating each robot's status are calculated by Equations \ref{eq:12} and \ref{eq:17}. 


\section{Experiment}
In order to validate the effectiveness of \textit{\textbf{Trust-R}} in helping with the swarm's self-diagnose of faulty behaviors and therefore restore human trust, two task scenarios were designed to compare the accumulated error and human trust before and after using \textit{\textbf{Trust-R}}. In these comparisons, the algorithm baseline was the traditional averaged-update control law (equation 1), and the trust baseline was human subjective rating of trust levels.

\subsection{Environment Design}
A simulation environment was designed based on the \textbf{\textit{CRAImrs}} framework, which was developed using simulation software Gazebo, task-related swarm control laws, and trust models \cite{Quigley2009ROS, Koenig2004gazebo}, as shown in Figure \ref{illustration_1}.
\textbf{\textit{CRAImrs}} was used for simulating human supervisory multi-robot teaming. The environment size was 50m$\times$50m. There were six quad-rotor UAVs involved in the following task scenarios, including one failed robot with degraded motor issue. With this faulty-robot involved robot teaming, this paper aimed to assess the effectiveness of our control method on correcting swarm behaviors under the influence of faulty robots. 


\subsection{Task Scenario Design}
Two typical tasks covering essential elements of emergent task response were tested.
\textbf{\textit{Scenario 1: transit between area inspection tasks. Scenario 2: response to emergent target "car accident area"}}. In the first scenario, the UAV swarm finished "activity inspection" and leave for "parking lot inspection" at target area 1. This task transition frequently appeared in emergency response and was designed here to observe the process of 
\textit{\textbf{Trust-R}} correcting robot behaviors and restore human trust in the swarm. In the second scenario, the UAV swarm was initially assigned to inspect parking lot (target area 1) while re-directed to inspect "target 2: car accident area" which is more urgent. This sudden change in emergent task response imposed challenges on adjusting speed, spatial distribution, and the heading directions.

\begin{figure*}[t]
  \centering
 \includegraphics [width=0.8 \linewidth ]{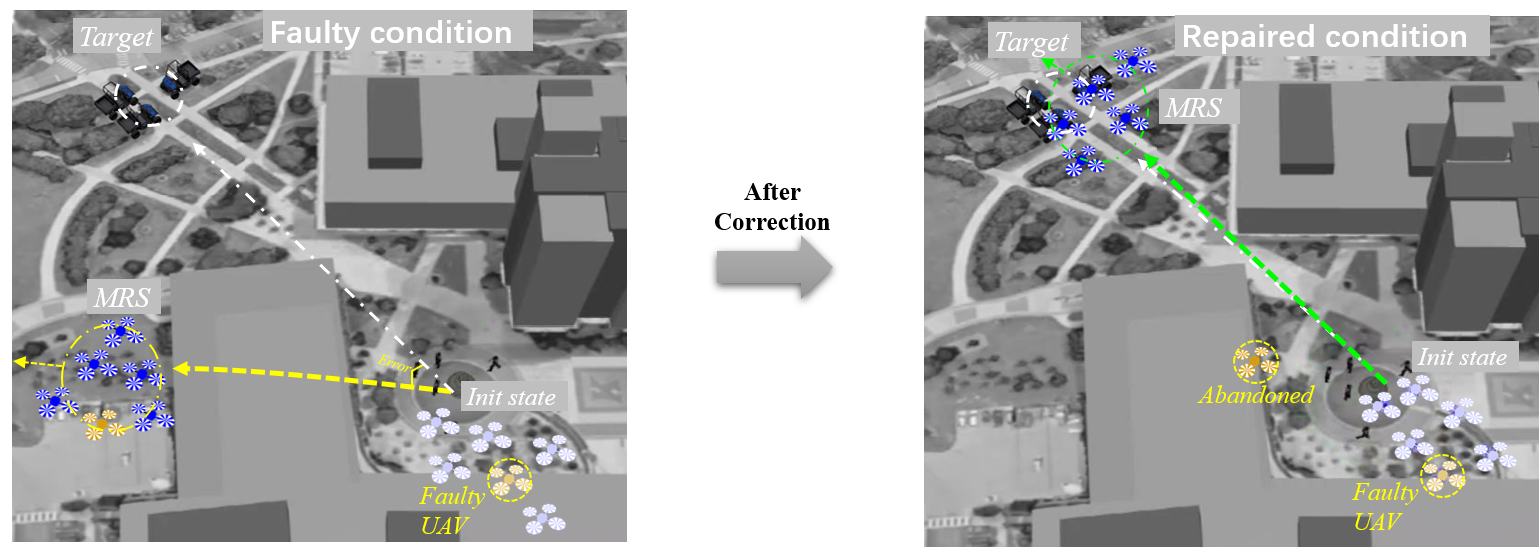}
  \caption{Experiment result for the Scenario: transitioning. While the UAV swarm deflects the expected trajectory in the faulty condition due to the presence of a failed robot, the UAV swarm with \textit{\textbf{Trust-R}} restricts the influence of the failed robot and maintains the expected trajectory to the target in the repaired condition.}
  \vspace{-1em}
  \label{Case-II}
\end{figure*}

\begin{figure*}[ht]
  \centering
 \includegraphics [width=0.8 \linewidth ]{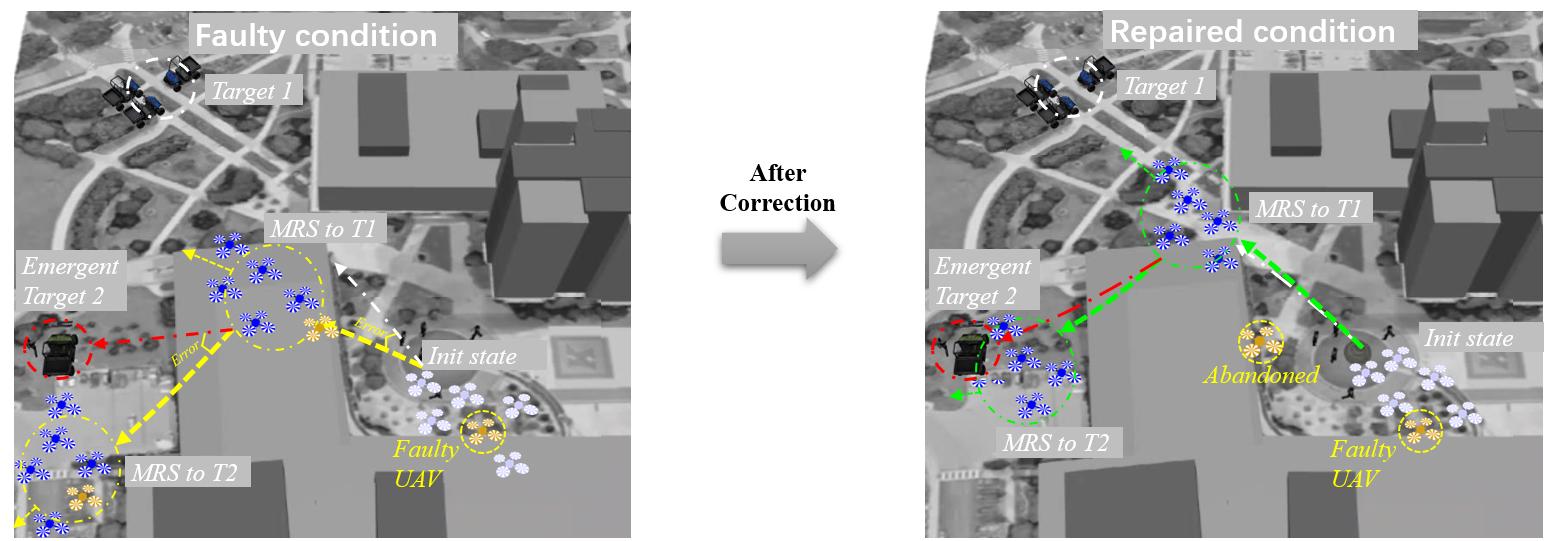}
  \caption{Experiment result for the scenario: emergency response. In the faulty condition, the UAV swarm with the failed robot deviates both Target 1 and Target 2. By constrast, the UAV swarm with \textit{\textbf{Trust-R}} can restrict the influence of the failed robot and flock to Targets 1 and 2 more precisely.}
  \vspace{-1em}
  \label{Case-III}
\end{figure*}

To further evaluate robustness of our method towards different errors, each task was designed with two levels of motor degradation. Motor degradation had a fixed offset perpendicular to the direction of velocity and a restricted maximum speed of 40\% and 70\%, respectively. The greater degradation (40\% speed) was expected to result in a more easily recognized abnormality. For each task, our method \textit{\textbf{Trust-R}} was applied to repair the faulty motion behaviors; while the control group used the traditional control method which is without a specially designed behavior repairing mechanism. Therefore there were totally eight task scenarios in our experiments.

\subsection{Human User Study} 

A human user study was conducted in the format of questionnaire published on the crowd-sourcing platform Amazon MTurk \cite{Buhrmester2016Amazon}. The full questionnaire can be found at this link \footnote{https://kent.qualtrics.com/jfe/form/SV\_0OImxiQRWOqxSId}. Totally there were 50 English-speaking volunteers recruited with a rate of US\$2 each. In order to guarantee data quality, volunteers were required to be Amazon Turk Masters and reach a minimum answer approval rate of 70\%. 

The user study comprised two main parts, a tutorial and experimental surveys collected after viewing each variant of the scenarios. In the experimental survey, the two task scenarios were presented sequentially; to avoid user training during the process, within each task scenario, the four conditions were randomized. In each trial, the participants monitored task progress and behaviors of the swarm, such as flocking speed, heading direction, and robot spatial relations (connectivity and formation). Participants were asked to observe the video of robot mission performance, and then decide whether or not a fault had occurred during the process. 
After that, a series of performance-related questions regarding specific UAVs were presented, including asking participants to decide whether a fault had occurred, identify robots showing faulty motion behaviors, and rate their trust levels on performance of both the swarm and specific robots. 


\section{Results}
Trust rating data was collected as ordinal categorical variables. The Mann-Whitney U test was used to analyze the effect of experimental factors on human trust. Human trust level was rated by five scales: \textit{Completely Distrust:1, Distrust:2, Neutral:3, Trust:4, Completely Trust:5}.

\textbf{\textit{Scenario I: task transitioning.}} 
Figure \ref{Case-II} shows the swarm performance with the control of our trust-weighted control method \textit{\textbf{Trust-R}} and the traditional averaged-update control method (Equation 1). The participants were more likely to report a fault in the scenarios of using traditional control method than that using \textit{\textbf{Trust-R}} (U = 1850, p $<$ .001), validating the effectiveness of our \textit{\textbf{Trust-R}} in repairing abnormal robot behaviors. 
The mean trust level for the faulty condition and repaired condition were 1.94 \textit{Distrust} and 3.83 \textit{Trust}, respectively; statistically, the participants were more likely to report a higher trust level in the repaired condition than the faulty condition (U = 2221, p $<$ .001). The trust rating result shows that the trust was effectively restored when faulty behavior was corrected.

In this scenario, the severity level of faults did not affect the proportion of human volunteers in reporting faulty behaviors (U = 1225, p $>$ .1).  This held both for participants in the faulty conditions (U = 1154,  p $>$ .1) and those in the repair conditions (U = 1177, p $>$ .1).
For the faulty condition, the participants showed $3.75$ average points of confidence (Very Difficult:1, Difficult:2, Neutral:3, Easy:4, Very Easy:5) in identifying failed robots among the swarm. In the faulty condition, 51\% of participants correctly identified the failed local collaboration containing the failed UAV; 11\% directly identified the failed robot. 

\textbf{\textit{Scenario II: emergency response}}. 
Figure \ref{Case-III} shows the results of the experiment for the scenario. 
The participants were more likely to report a fault in the faulty conditions than in the repaired conditions (U = 2250, p $<$ .001). 
This significant difference in identifying faulty behaviors shows that in the emergent motion adjustment, the team performance got degraded seriously by the robot faulty behaviors and validated that our \textit{\textbf{Trust-R}} method is effective in repairing swarm behaviors. The mean trust level for the faulty condition and repaired condition were Distrust (2.18) and Trust (4.08), respectively; the participants were more likely to report a higher trust level in the repaired condition than the faulty condition (U = 1985, p $<$ .001), showing that the trust could be restored when faulty behavior was corrected.

In the scenario the severity of faults affected the chance that they were reported (U = 1000, p = 0.013). In the faulty condition, the effects of severity extended to trust rating (U = 965, p = 0.019). However, in the repaired condition, rated trust was unaffected (U = 1238, p $>$ .1).
For the faulty condition, the participants rated their identification of failed robots as \textit{Easy} ($3.99$). In the faulty condition, 41\% of participants correctly identified the failed UAV by observing abnormal local cooperation, while 17\% of the participants identified the failed robot by observing a single-robot's motion behaviors. 

Table \ref{tab:dynamic_scenario} presents the human trust levels of the faulty and repaired conditions for UAV swarm for the scenarios. Our method generally improved the trust level of a swarm from 2.0 (5 is full scale) to 4, validating the effectiveness of our method in restoring trust on swarm.
Table \ref{tab:Error} summarizes the swarm behaviors under the control of the traditional method and the \textit{\textbf{Trust-R}} method. Results showed the traditional method had a poor fault resiliency as the swarm's head direction deviated 20-40 degrees during emergency response; after \textit{\textbf{Trust-R}} correction, the direction was corrected to the original plan with only 3-4 degrees deviation. After \textit{\textbf{Trust-R}} correction, the distance to target area was limited within 3 meters while there are 10-20 meters deviations using the traditional control method. The better mission performance with the \textit{\textbf{Trust-R}} method showed the effectiveness of \textit{\textbf{Trust-R}} in correcting swarm behaviors and ensuring a high-quality emergency response.

\begin{table}[h]
\caption{emergent scenario conditions \\ $^\ast$ Mann-Whitney U}
\label{tab:dynamic_scenario}
\begin{center}
\begin{tabularx}{0.45\textwidth} {
   >{\centering\arraybackslash}X 
  | >{\centering\arraybackslash}X 
   >{\centering\arraybackslash}X 
   >{\centering\arraybackslash}X}
 \hline
 \hline
 \multirow{2}{8em}{Swarm Status} & \multicolumn{3}{>{\hsize=\dimexpr3\hsize+2\tabcolsep+\arrayrulewidth\relax\centering\arraybackslash}X}{Median Trust Level}\\
 \cline{2-4} 
   & Faulty  & Repaired  & $p^\ast$ \\
 \hline
 Scenario 1 & Distrust/1.94  & Trust/3.83  & $<$0.001\\
 \hline
 Scenario 2 & Distrust/2.18  & Trust/4.08  & $<$0.001\\
 \hline
 \hline
\end{tabularx}
\end{center}
\end{table}

\begin{table}[h]
\caption{Values of flocking heading-direction and final distance to target}
\label{tab:Error}
\begin{center}
\begin{tabularx}{0.5\textwidth} {
    >{\centering\arraybackslash}X 
  | >{\centering\arraybackslash}X 
   >{\centering\arraybackslash}X 
   >{\centering\arraybackslash}X 
  | >{\centering\arraybackslash}X 
   >{\centering\arraybackslash}X 
    >{\centering\arraybackslash}X}
 \hline
 \hline
 \multirow{2}{5em}{scenario}  & \multicolumn{3}{>{\hsize=\dimexpr3\hsize+2\tabcolsep+\arrayrulewidth\relax\centering\arraybackslash}X|}{heading-direction/$^o$} & \multicolumn{3}{>{\hsize=\dimexpr3\hsize+2\tabcolsep+\arrayrulewidth\relax\centering\arraybackslash}X}{distance to target/m}\\
 \cline{2-7} 
    & Designed & Faulty & Repaired & Designed & Faulty & Repaired\\
 \hline
  1 & 43 & 3  & 46 & 0.0 & 20.6 & 2.0\\
 \hline
  2 & 0 & -19  & -4 & 0.0 & 9.8 & 3.2\\
 \hline
 \hline
\end{tabularx}
\end{center}
\end{table}

\section{Conclusion$\&$Future Work}

Our previous research of \textit{\textbf{Trust-R}} investigated the general control method of mitigating failure influence on a robotic swarm, which was supervised by a human \cite{liu2019trust, liu2019trustRepair}. This research, based on a real-gravity simulation environment with physical robot models, continuously investigated the effectiveness of using the \textit{\textbf{Trust-R}} method for emergency response, which required a swarm to immediately disengage the current undergoing task for a more urgent task. This fundamental research has a huge impact on robot-assisted natural disaster rescue, which dynamically pops up more urgent tasks and requires immediate robot assistance. With typical scenarios "transit between area inspection tasks" and "response to emergent target -- car accident", the effectiveness of \textit{\textbf{Trust-R}} in supporting emergency response was validated. This research delivered a simulation pipeline that integrated human user study, robot simulation platform and control algorithms, providing a guideline for future human-supervisory MRS research. The protocols used in our user trust study are also valuable for future trustworthy MRS research.

In the future, more fault factors caused by unstable systems and environmental disturbances will be considered so that a more accurate trust-based method can be designed to improve the performance of a UAV swarm for emergent tasks.

\end{document}